\documentclass[letterpaper, 10 pt, conference]{ieeeconf}  

\IEEEoverridecommandlockouts                              

\usepackage{graphics}       
\usepackage{epsfig}         
\usepackage{times}          
\usepackage{amsmath}        
\usepackage{amssymb}        
\usepackage{afterpage}
\usepackage{makecell}
\usepackage{tabularx}
\usepackage{algorithmic}
\usepackage{caption}
\usepackage{array}
\usepackage{adjustbox}
\usepackage[utf8]{inputenc}
\usepackage[normalem]{ulem}
\usepackage{multicol}

\usepackage{amsmath,amssymb} 
\usepackage{graphicx}
\usepackage{textcomp}
\usepackage{epsfig} 
\usepackage{subfigure}
\usepackage{capt-of}
\usepackage{amsmath} 
\usepackage{amssymb}  
\usepackage{epstopdf}
\usepackage{lipsum}
\usepackage{lipsum,amsmath,multicol}
\usepackage{float}
\usepackage[linesnumbered,ruled,vlined]{algorithm2e}
\usepackage{algorithm2e}

\SetCommentSty{mycommfont}

\SetKwInput{KwInput}{Input}        
\SetKwInput{KwReturn1}{Return}    
\SetKwInput{KwOutput}{Output} 
\SetKwInput{KwInitialization}{Initialization} 
\usepackage{wrapfig}
\usepackage{sidecap}

\usepackage{amsthm}

\usepackage{comment}
\usepackage{array}
\usepackage{glossaries}
\usepackage{breqn}
\usepackage{mathtools, cuted}
\makeglossaries
\newcolumntype{L}[1]{>{\raggedright\let\newline\\\arraybackslash\hspace{0pt}}m{#1}}
\newcolumntype{C}[1]{>{\centering\let\newline\\\arraybackslash\hspace{0pt}}m{#1}}
\newcolumntype{R}[1]{>{\raggedleft\let\newline\\\arraybackslash\hspace{0pt}}m{#1}}
\usepackage{sidecap}
\usepackage{url}
\usepackage[T1]{fontenc}
\usepackage{tabularx}
\usepackage{multicol, blindtext}
\usepackage{amsmath}
\usepackage{multirow}

\usepackage[table,xcdraw]{xcolor}
\usepackage{epsfig}         
\usepackage{times}          
\usepackage{amsmath}        
\usepackage{amssymb}        
\usepackage{afterpage}
\usepackage{makecell}
\usepackage{tabularx}
\usepackage{algorithmic}
\usepackage{caption}
\usepackage{array}
\usepackage{adjustbox}
\usepackage[utf8]{inputenc}
\usepackage[normalem]{ulem}
\usepackage{multicol}
\usepackage{hyperref}
\usepackage{cleveref}
\usepackage[style=ieee]{biblatex} 
\def\equationautorefname~#1\null{(eq~#1)\null}
\def\subsectionautorefname~#1\null{(sec~#1)\null}

\usepackage{mathtools, bigstrut}
\usepackage{tabularx}
\usepackage{multicol, blindtext}
\usepackage{amsmath}
\usepackage{multirow}

\allowdisplaybreaks

\usepackage{mathtools, bigstrut}
\usepackage{tabularx}
\usepackage{multicol, blindtext}
\usepackage{amsmath}
\usepackage{multirow}

\def\equationautorefname~#1\null{eq~(#1)\null}
\def\subsectionautorefname~#1\null{sec~(#1)\null}

\title{\LARGE \bf
EDMP: \underline{E}nsemble-of-costs-guided \underline{D}iffusion for \underline{M}otion \underline{P}lanning 
}

\author{Kallol Saha$^{*1}$, Vishal Mandadi$^{*1}$, Jayaram Reddy$^{*1}$, Ajit Srikanth$^{1}$, \\Aditya Agarwal$^{2}$, Bipasha Sen$^{2}$, Arun Singh$^{3}$, and Madhava Krishna$^{1}$ 
\thanks{${}^*$Denotes equal contribution}
\thanks{Project website \& codebase: \href{https://ensemble-of-costs-diffusion.github.io/}{https://ensemble-of-costs-diffusion.github.io/}}
\\
{\small $^{1}$Robotic Research Center, IIIT Hyderabad,}
{\small $^{2}$Massachusetts Institute of Technology,}
{\small $^{3}$University of Tartu}
}

\begin{document}

\makeatletter
\let\@oldmaketitle\@maketitle
\renewcommand{\@maketitle}{\@oldmaketitle
\centering 
\includegraphics[width=\linewidth]{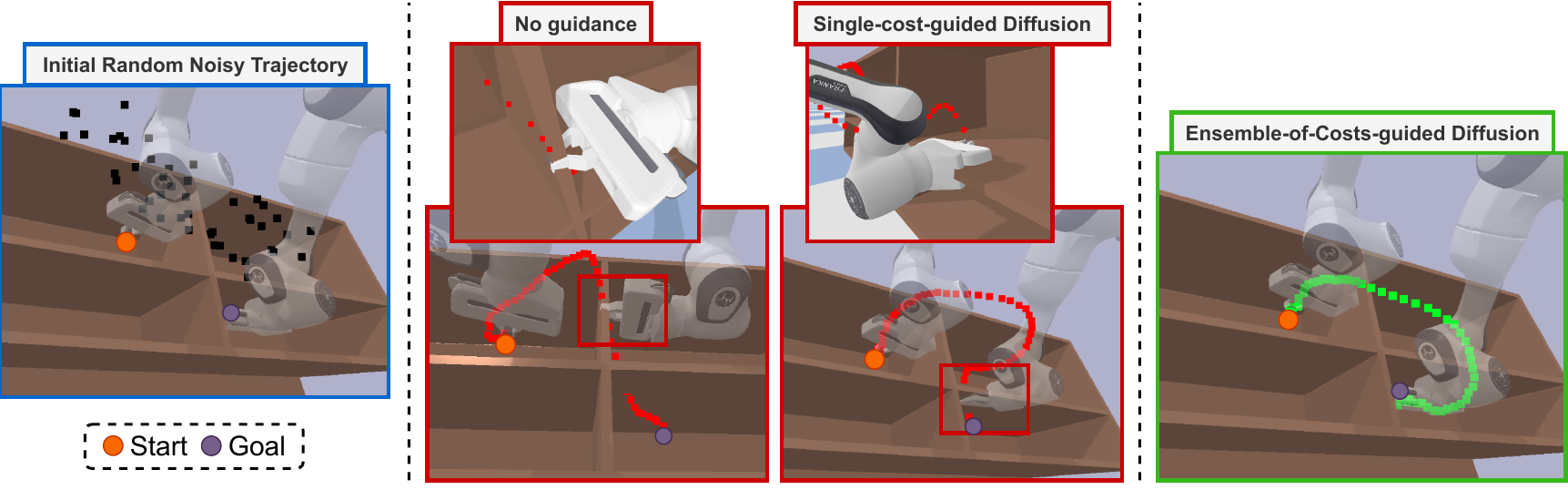}
\captionof{figure}{\small \textbf{E}nsemble-of-costs-guided \textbf{D}iffusion for \textbf{M}otion \textbf{P}lanning combines the strength of classical planning and deep-learning. We leverage a diffusion policy to learn a prior over kinematically \textit{valid} trajectories and guide it directly at the time of inference using scene-specific costs such as "collision-cost" to 
also satisfy the scene-specific constraints. Instead of using a single-cost, we propose the use of multiple cost functions (ensemble-of-cost-guidance) to capture variations across scenes, enabling us to generalize to diverse scenes.}
\label{fig:teaser}
\vspace{-10px}
}
\makeatother

\maketitle
\thispagestyle{empty}
\pagestyle{empty}


\begin{abstract}
Classical motion planning for robotic manipulation includes a set of general algorithms that aim to minimize a \textit{scene-specific cost} of executing a given plan.
This approach offers remarkable adaptability, as they can be directly used off-the-shelf for any new scene without needing specific training datasets. 
However, without a prior understanding of what diverse valid trajectories are and without specially designed cost functions for a given scene, the overall solutions tend to have low success rates. While deep-learning-based algorithms tremendously improve success rates, they are much harder to adopt without specialized training datasets. We propose EDMP, an \underline{E}nsemble-of-costs-guided \underline{D}iffusion for \underline{M}otion \underline{P}lanning that aims to combine the strengths of classical and deep-learning-based motion planning. Our diffusion-based network is trained on a set of diverse kinematically valid trajectories. 
Like classical planning, for any new scene at the time of inference, we compute scene-specific costs such as "collision cost" and guide the diffusion to generate valid trajectories that satisfy the scene-specific constraints. 
Further, instead of a single cost function that may be insufficient in capturing diversity across scenes, we use an \textit{ensemble} of costs to guide the diffusion process, significantly improving the success rate compared to classical planners. 
EDMP performs comparably with SOTA deep-learning-based methods while retaining the generalization capabilities primarily associated with classical planners.

\end{abstract}


\section{INTRODUCTION}

Planning a trajectory from a start to goal position while avoiding collisions (self-collisions and collisions with the objects around the robot) is a fundamental challenge in robotic manipulation~\cite{agarwal2022approaches}. Over the years, many approaches have been introduced to tackle this challenge, including classical planning algorithms like ~\cite{5152817_ratliff, bhardwaj2022storm, inproceedings_schulman_ho_jonathan_lee, 4082128, inproceedings, conf/nips/LikhachevGT03, Kuffner2000RRTconnectAE, LaValle1998RapidlyexploringRT, inproceedings_bangura, 7029990_erez, Williams2015ModelPP, Jankowski2022VPSTOVS, 7487277}, and more recent deep-learning-based algorithms like MPNets~\cite{8793889_qureshi}, and M$\pi$Nets~\cite{fishman2023motion}. While the latter approaches tremendously improved the success rate in the manipulation tasks (determined by the number of times the manipulator reaches the goal while avoiding collisions), classical planners greatly reduce the dependency on specialized training datasets, generalizing to novel scenes and therefore are still widely the go-to off-the-shelf choice for motion planning. 

In a classical planning approach, a motion plan for a new scene is generated on the go by minimizing a \textit{cost function} computed on the given scene. For example, 
minimizing collision cost, path length, and the distance to goal. 
However, designing this cost function with multiple constraints requires many hit-and-trials, and careful fine-tuning, and still may not capture the diversity \textit{across} scene structures. 



Deep learning-based methods~\cite{danielczuk2021object} improve upon this by adopting data-driven techniques that learn a mapping from a given context (or the scene) to a solution using a neural network via behavior cloning~\cite{osa2018algorithmic}. This allows the network to gain an overall understanding of different scene structures and map the structures to a viable solution. 
Such methods are significantly faster and more accurate. Despite the gains, they falter in generalization -- performance on out-of-distribution scenes is significantly impacted. They also require a large number of high-quality demonstrations and are shown to be inefficient when fed with multimodal datasets~\cite{pearce2023imitating}, for example, data collected using teleoperation~\cite{zhao2023learning_aloha}. 

In this work, we aim to bridge the gap between the two approaches by first learning a prior (as in deep-learning-based methods) over kinematically 
\textit{valid }trajectory for \textit{any} scene and then incorporating \textit{scene-specific cost}, such as collision cost, directly at the time of inference (as in classical planners). 
As shown in the experimental section, this builds a powerful motion planner that generalizes to diverse scenes. 


Recently, diffusion models~\cite{sohl2015deep, ho2020denoising} have gained tremendous popularity for generating diverse sets of modalities such as vision~\cite{dalle, dalle2, imagen}, language~\cite{audio_video_diffusion}, and more recently in motion planning~\cite{janner2022planning, chi2023diffusion, liang2023adaptdiffuser, ajay2022conditional}. Unlike generative models like GANs~\cite{goodfellow2014generative} and VAEs~\cite{kingma2022autoencoding}, diffusion models generate datapoints in multiple timesteps. In the backward pass (generation step), a diffusion process generates the output by iteratively removing "noise" in $h$ steps to eventually denoise a given noisy input. For example, generating a kinematically valid trajectory from a noisy trajectory. As shown in~\cite{janner2022planning, carvalho2023motion}, this allows for a unique opportunity to "guide" the generation process by incorporating "scene-specific cues," (such as a collision cost) as in the classical planners, at each of the intermediate steps, thereby producing a kinematically valid trajectory that also satisfies the scene-specific constraints.
Taking inspiration from this, we propose, \textbf{EDMP}, an \textbf{E}nsemble-of-costs-guided \textbf{D}iffusion for \textbf{M}otion \textbf{P}lanning, that first learns a prior of kinematically valid trajectories and then is guided at the time of inference to satisfy scene-specific constraints while retaining the kinematic validity. Further, instead of relying on a single cost function, we incorporate  $N$ different cost functions (ensemble-of-costs) to guide the diffusion process, thereby generating trajectories that can solve diverse challenging scenes (as shown in Fig~\ref{fig:teaser}).

\begin{figure}
    \centering
    \includegraphics[width=\linewidth]{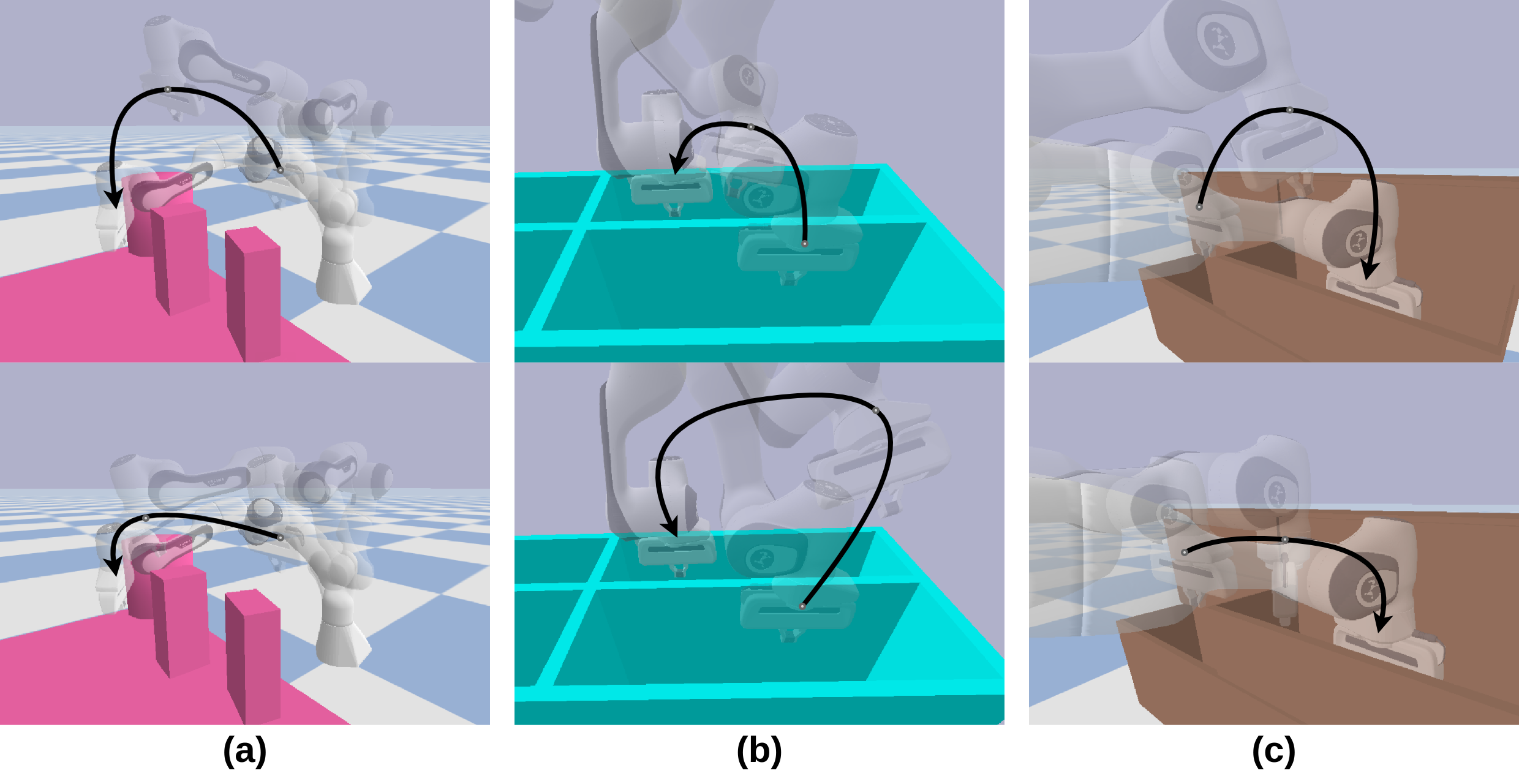}
    \vspace{-10px}
    \caption{\small \textbf{Multimodal trajectory predictions in } (a) tabletop scene, (b) merged cubby scene, and (c) dresser scene.}
    \label{fig:mm}
    \vspace{-13px}
\end{figure}

Interestingly, based on the concepts of diffusion, EDMP can generate multimodal trajectories (multiple ways of getting from point A to point B). The ensemble-of-costs further improves the diversity in multimodality 
that could be handpicked on different parameters such as path length or smoothness (see Figure~\ref{fig:mm}).  Our key contributions are:

\begin{enumerate}
    \item We propose \textbf{EDMP}, an \textbf{E}nsemble-of-costs-guided \textbf{D}iffusion for \textbf{M}otion \textbf{P}lanning
    which combines the strengths of classical and deep learning-based motion planning by first learning a prior over kinematically valid trajectories and then using multiple cost functions to capture diverse scene-specific cost guidance directly at the time of inference. 
    \item EDMP generalizes to diverse novel  
    and even out-of-distribution scenes (such as planning for a manipulator holding an arbitrary object) while performing comparably with SOTA deep-learning-based methods on specific datasets. 
    \item Based on diffusion, EDMP generates multimodal trajectories. Further, the ensemble-of-costs improves the diversity of multimodality. This can be exploited by downstream planners to choose a trajectory from the set of solutions that evaluates to an optimal objective.

\end{enumerate}


\section{Problem Formulation}


Consider a robotic manipulator with $m$ joints corresponding to a given joint state $\boldsymbol{s}_i \in \mathbb{R}^m$ for the $i$\textsuperscript{th} trajectory-time-step where a trajectory, $\boldsymbol{\tau}$, is a sequence of joint states over a horizon $h$ between a start and goal configuration, $\boldsymbol{s}_0$ and $\boldsymbol{s}_{h-1}$, respectively, given as, $\boldsymbol{\tau} = [s_0,\space s_1, \space ... \space \ \boldsymbol{s}_{h-1}]^\top \in \mathbb{R}^{m\times h}$. We are interested in solving the following optimization problem for obtaining a trajectory for a given task,
\begin{align}
    \min_{\boldsymbol{\tau}} J(\boldsymbol{\tau}; \textbf{o}, \boldsymbol{E})
    \label{traj_opt_problem}
\end{align}

The vector $\textbf{o}$ represents the hyperparameters of the cost function $J$. The variable $\boldsymbol{E}$ represents the scene. The cost function has two distinct parts.  The first part models aspects that depend purely on the manipulator's kinematics, such as smoothness and feasibility (e.g. self-collision) of the joint trajectory. The second part represents the scene-specific cost that couples the manipulator motion with the scene. 

\subsection{Diffusion Based Optimal Solution}
Diffusion models~\cite{sohl2015deep, ho2020denoising} are a category of generative models where a datapoint is converted into an isotropic Gaussian noise by iteratively adding Gaussian noise through a fixed forward diffusion process $q(\boldsymbol{\tau}_t|\boldsymbol{\tau}_{t-1}) = \mathcal{N}(\boldsymbol{\tau}_t; \sqrt{1 - \boldsymbol{\beta}_t}\boldsymbol{\tau}_{t-1}, \boldsymbol{\beta}_t \text{I})$, where $\boldsymbol{\beta}_t$ is the variance schedule and $t$ is the diffusion timestep. 
A forward diffusion process of $T$ timesteps can be reversed by sampling $\boldsymbol{\tau}_T \sim \mathcal{N}(0, \text{I})$ from a standard normal distribution and iteratively removing gaussian noise using the trainable reverse process $p_{\theta}(\boldsymbol{\tau}_{t-1}/\boldsymbol{\tau}_t) = \mathcal{N}(\boldsymbol{\tau}_{t-1}; \mu_{\theta}(\boldsymbol{\tau}_t, t), \boldsymbol{\Sigma}_t)$ parametrized by $\theta$.

\begin{figure*}[t!]
    \centering
    \includegraphics[width=0.85\linewidth]{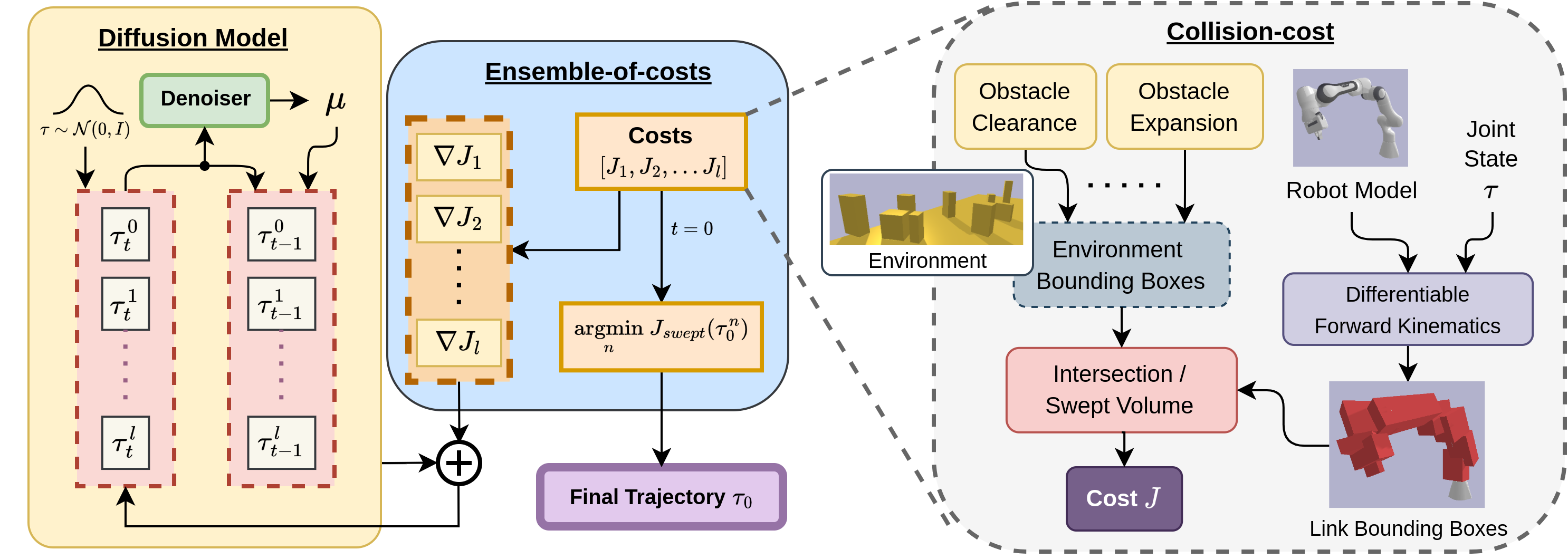}
    \caption{\small \textbf{Architecture}. EDMP leverages a diffusion model alongside an ensemble of $l$ cost functions. The diffusion model denoises a batch of trajectories $\boldsymbol{\tau}$ from $t = T$ to $0$, while each cost in the ensemble guides a specific sub-batch. We calculate the gradient $\nabla J$ of each collision cost (intersection or swept volume) in the ensemble from robot and environment bounding boxes, using differentiable forward kinematics. After denoising is over, the trajectory with minimum swept volume is chosen as the solution.}
    \label{fig:architecture}
    \vspace{-10px}
\end{figure*}

We train a diffusion model to learn a prior $p_\theta$, over a set of smooth kinematically valid trajectories collected from a large-scale dataset~\cite{fishman2023motion}. For a given trajectory in the dataset, forward diffusion first corrupts it by gradually adding noise, and then the reverse diffusion process aims to reconstruct back the original trajectory from the noise. This is shown in Figure~\ref{fig:architecture}-(Diffusion Model). 


\textbf{Conditioning on start and goal: }As we train the diffusion prior on the set of trajectories from the dataset, we 
repeatedly fix $\boldsymbol{s}_0$ and $\boldsymbol{s}_{h-1}$ at each diffusion-time-step to correspond to the original $\boldsymbol{s}_0$ and $\boldsymbol{s}_{h-1}$ at the first diffusion-time-step for each trajectory. As shown in Section~\ref{sec:ablations}, this creates a conditioning effect on the start-goal pair, encouraging the network to generate a smooth trajectory between the two.



\subsection{Guidance Process}



\begin{figure}[]
    \centering
    \includegraphics[width=\linewidth]{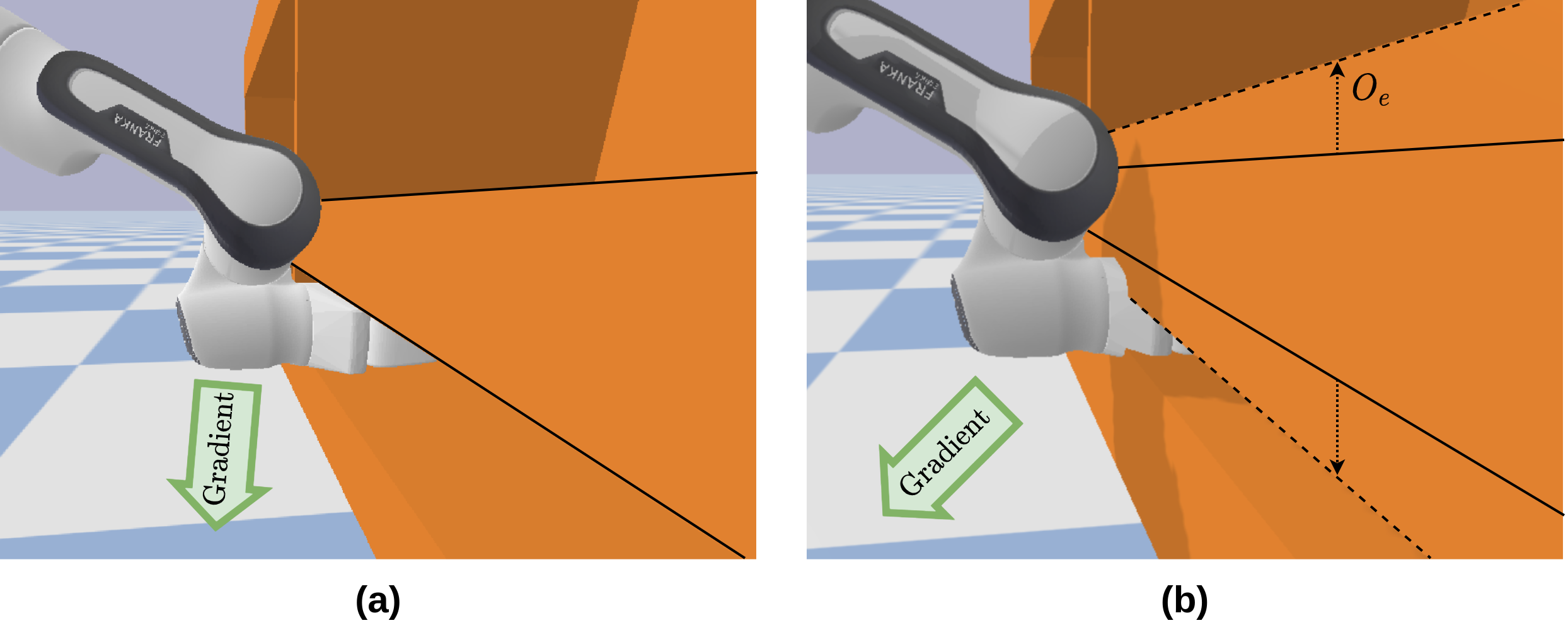}
    \caption{\small \textbf{Effect of Obstacle Expansion.} (a) scene without obstacle expansion applied. (b) shows the change in the scene after applying obstacle expansion. Obstacle Expansion widens the thinner dimension by an amount $O_e$, thus altering the gradient direction enabling the manipulator find a configuration for retracting around the shelf.}
    \label{fig:expansion}
    \vspace{-15px}
\end{figure}
For a specific scene, we also want to incorporate a \textit{scene-specific cost} function that can guide the diffusion process toward a trajectory that satisfies the scene-specific constraints, such as collision cost, denoted by $J(\boldsymbol{\tau}; \textbf{o}, \boldsymbol{E})$.
Specifically, at each step of the denoising-time-step, we modify the intermediate trajectory $\boldsymbol{\tau}_t$ predicted by the denoiser at the $t$\textsuperscript{th} diffusion-time-step by adding gradients from the scene-specific collision cost function, before passing to the next step as shown in Figure~\ref{fig:architecture}. This is given by, 
\begin{equation}
    \boldsymbol{\tau}^*_t= \boldsymbol{\tau}_t - \alpha\nabla J(\boldsymbol{\tau}; \textbf{o}, \boldsymbol{E})
    \label{eqn:guidance}
\end{equation}
where $\alpha$ is a hyperparameter. This form of conditioning is analogous to classifier-based guidance in~\cite{dhariwal2021diffusion}, where a trained classifier guides the diffusion towards a goal. In our case, it guides the trajectory to collision-free regions. Although adding cost-conditioning directly at the time of inference results in a cost-guided posterior, denoising from the cost-guided posterior is equivalent to denoising from a Gaussian distribution, as shown in~\cite{carvalho2023motion}. 

\section{Ensemble-of-Costs guided Diffusion}

\label{sec:ensemble}
Equation~\ref{eqn:guidance} uses gradients from a scene-specific cost function to modify the trajectory obtained from the diffusion prior. 
Thus, the nature of the gradient dictates the efficacy of the process, which in turn depends on many factors, some of which we summarize below:

\begin{itemize}
    \item \textbf{Algebraic form of the cost model:} Collision cost is highly scene-specific. For example, the signed-distance-field-based collision cost~\cite{5152817_ratliff} is known to struggle in scenes with narrow racks between the start and the goal positions. Similarly, performance of collision costs modeled around convex collision checking~\cite{inproceedings_schulman_ho_jonathan_lee} depends on whether we penalize each independent waypoint's intersection volume with the environment (intersection volume) or the swept volume between each of the adjacent waypoints (swept volume) as shown in Figure~\ref{fig:iv_and_sv}.

\item \textbf{Gradient and Cost Hyperparameters:} Each collision cost model has several hyperparameters that also affect the gradient descent step in Equation~\ref{eqn:guidance}. For example, gradient weight schedule, whether adaptive or independent, is an important parameter in gradient-based approaches. Similarly, we show that in some cases, choosing to expand the object over the initial period of optimization greatly helps in bringing out the retraction behavior, which otherwise seemed less likely as shown in Figure~\ref{fig:expansion} (retraction prevents the manipulator from colliding with the shelf).

\end{itemize}

The above discussion makes the case for guided diffusion with an adaptive learning rate and collision cost whose algebraic form and hyper-parameters are adaptive to the environment in which the manipulator is operating. However, to the best of our knowledge, there exists no general framework towards this end. 
We propose an ensemble-of-collision-costs where we run a batch of guided diffusion processes in parallel, as shown in Figure ~\ref{fig:architecture} where each sub-batch uses a different cost function. Moreover, we embed the hyperparameter tuning of the cost functions as a part of the diffusion process. Algorithm\ref{algo2} summarizes our proposed improvements.

We assume that we have $l$ different choices of the cost function stemming from different collision models with each having their own hyperparameters. We also consider $r$ different schedules to adapt the hyperparameter of each of the $l$ costs. Thus, in Algorithm~\ref{algo2}, we have a triple loop that runs $l\times r$ parallel diffusion processes. In practice, however, the outer two loops are decoupled from each other and run in batch fashion over GPUs. 

\begin{algorithm}[t]
\DontPrintSemicolon
\caption{Reverse Diffusion Guided with a Ensemble of Costs}\label{algo2}
\KwInput{Learned reverse diffusion $\boldsymbol{\mu}_{\theta}(\boldsymbol{\tau}_t, t)$, Covariance schedule $\boldsymbol{\Sigma}_t$, learning rate schedule $\alpha_t$, hyperparameter schedule $\mathbf{o}_t$ }
\KwInitialization{$\boldsymbol{\tau}_T \sim \mathcal{N} (\textbf{0}, \textbf{I})$, Start joint state $\boldsymbol{s}_0$, final joint state $\boldsymbol{s}_{h-1}$}
\For{$j = 1, \dots, \dots l$}{
\For{$i = 1, \dots, \dots r$}{
\For{$t = T, \dots, \dots 1$}{
   ${^{j, i}}\boldsymbol{\tau}_{t-1} \sim \mathcal{N}(\boldsymbol{\mu}_{\theta}( {^{j, i}} \boldsymbol{\tau}_t, t )+{^{j, i}}\alpha_t \nabla_{\boldsymbol{\tau}} {^{j, i}}J(\mathbf{o}_t), \boldsymbol{\Sigma}_t  )$ 
}
}
}
\textbf{Return} ${^j}\boldsymbol{\tau}_0$ corresponding to minimum collision cost.
\label{algo:reverse_diff}
\end{algorithm}

\subsection{Collision Cost Guidance}
\label{sec:cost_guidance}

\begin{figure}[t]
    \centering
    \includegraphics[width=\linewidth]{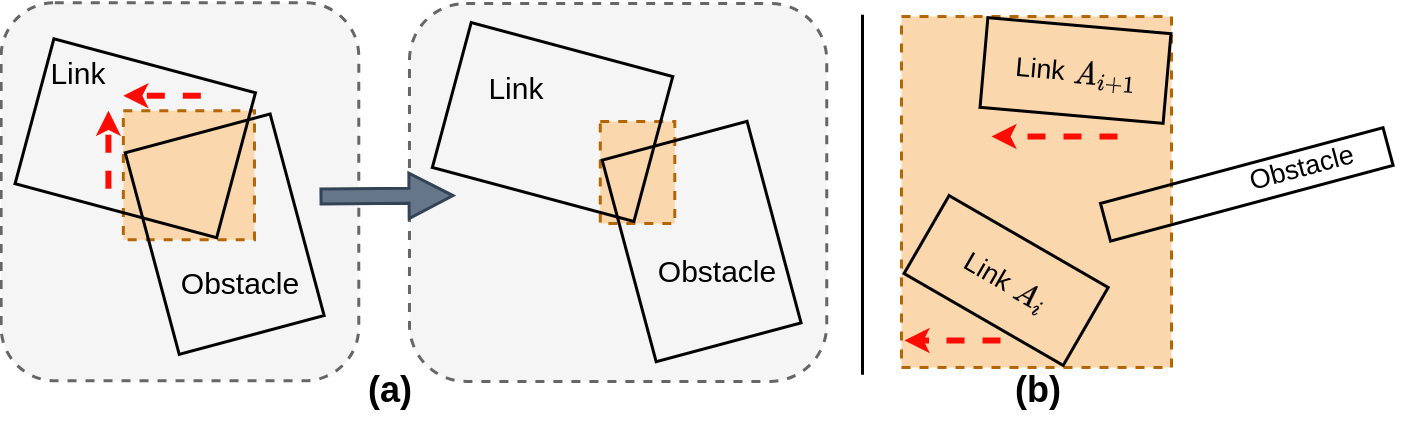}
    \caption{\small \textbf{Intersection v/s Swept Volume}. (a) Gradient (red arrows) of link-obstacle intersection volume moves link away from obstacle. (b) Gradient (red arrows) of swept volume between consecutive link poses prevents collision between trajectory waypoints}
    \label{fig:iv_and_sv}
    \vspace{-15px}
\end{figure}

To detect and compute the collision cost between any two rigid bodies, we take inspiration from Gilbert-Johnson-Keerthi (GJK)~\cite{2083} algorithm and the Expanding Polytope Algorithm (EPA)~\cite{article}. We compute the collision cost between two bodies, $A$ and $B$ as a function of overlap between them along the three axes. We refer to this as the penetration depth. To achieve this, we enclose each object in a bounding box cuboid and compute the intersection volume $V$ as:
\begin{equation}
\begin{array}{cc}
   \text{V}(A, B) &= \text{prod}(|\max(\min(A),\  \min(B))
     \\ & - \min(\max(A),\ \max(B))|) 
\end{array}
\end{equation}


To leverage this in manipulator collision avoidance, we first approximate each object on the table as a cuboid corresponding to a 3D bounding box that encloses the full object. Similarly, we represent each manipulator link as a 3D cuboid, which acts as a bounding box enclosing that link. We compute the poses of each of these bounding boxes for a given configuration $s_k$ using differentiable forward kinematics $\text{FK}(\boldsymbol{s}_k[i])$, where $\boldsymbol{s}_k[i]$ denotes the configuration of $i^{th}$ link. If there are $n$ obstacles in the scene, we can approximate the scene as a list of $n$ bounding boxes $\boldsymbol{E}\in \mathbb{R}^{n \times 4 \times 4}$. Each bounding box is represented as a transformation matrix of size $4 \times 4$ in the SE(3) Lie Group. We then compute the collision cost from the intersection volume, which we call "intersection volume cost", $J_{inter}$, of each link state in $\boldsymbol{\tau}$ with respect to each obstacle in $E$ given as:
\begin{equation}
    J_{inter}(\boldsymbol{\tau}, \boldsymbol{E}) = \sum_{k=0}^{h-1}\sum_{i=0}^{m-1} \text{V}(\text{FK}(\boldsymbol{s}_k[i]), \boldsymbol{E})
\label{eqn:intersection}
\end{equation}

%

The gradient of intersection volume cost $\nabla J_{inter}$ gives the penetration depth along each dimension. Averaging these values produces an average movement direction for a link along the penetration depth, to displace it out of collision, as shown in Figure~\ref{fig:iv_and_sv}. We show empirically that this simple collision cost enables collision avoidance even in complex scenes. 
 
 We define a swept volume cost function from swept volume $SV$, inspired from~\cite{2083}, as:
\begin{equation}
\begin{array}{cc}
   SV(\boldsymbol{s}_k[i], \boldsymbol{s}_k[i+1]) = 
   \\ \text{prod}(|\max(\text{FK}(\boldsymbol{s}_k[i]), \text{FK}(\boldsymbol{s}_k[i+1])) 
     \\ - \min(\text{FK}(\boldsymbol{s}_k[i]), \text{FK}(\boldsymbol{s}_k[i+1]))|)
\end{array}
\end{equation}

The swept volume approximates the volume swept out by a link between two adjacent waypoints. Such a cost function helps account for collisions that may happen between consecutive joint states in a trajectory, as in Figure~\ref{fig:iv_and_sv}. 

For a given trajectory $\boldsymbol{\tau}$, and environment $\boldsymbol{E}$, we define swept-volume cost as the sum of volumes swept out by the links between each pair of adjacent waypoints. As the swept volume is also modelled as a cuboid, we can plug it into Equation~\ref{eqn:intersection} to get the swept volume cost:
\begin{equation}
\label{eqn:swept_vol_cost}
J_{swept}(\boldsymbol{\tau}, \boldsymbol{E}) = \sum_{i=0}^{h-2}\sum_{i=0}^{m-1} \text{V}(\text{SV}(\boldsymbol{s}_k[i], \boldsymbol{s}_k[i+1]), \boldsymbol{E})
\end{equation}

Our architecture is also outlined in Figure~\ref{fig:architecture}.

\label{collision_cost}

\section{EXPERIMENTS}

All our experiments are conducted using the Pybullet simulator~\cite{coumans2021}, employing the Franka Panda robot.

\textbf{Dataset: } We benchmark on the M$\pi$Nets~\cite{fishman2023motion} dataset. 

The \textbf{training} dataset consists of 6.54 million collision-free trajectories. These 6.54 million trajectories are generated on various scenes (such as tabletop, dresser, and cubby) using two different classical planning pipelines: \textbf{Global} Planner (3.7 million trajectories) based on AIT$^*$~\cite{strub2020adaptively} and \textbf{Hybrid} Planner (3.7 million trajectories) that combined AIT$^*$~\cite{strub2020adaptively} for planning in the end-effector space and Geometric Fabrics~\cite{xie2021geometric} for producing a geometrically consistent motion conditioned on the generated end-effector waypoints. 

\begin{figure*}
    \centering
    \includegraphics[width=0.9\textwidth]{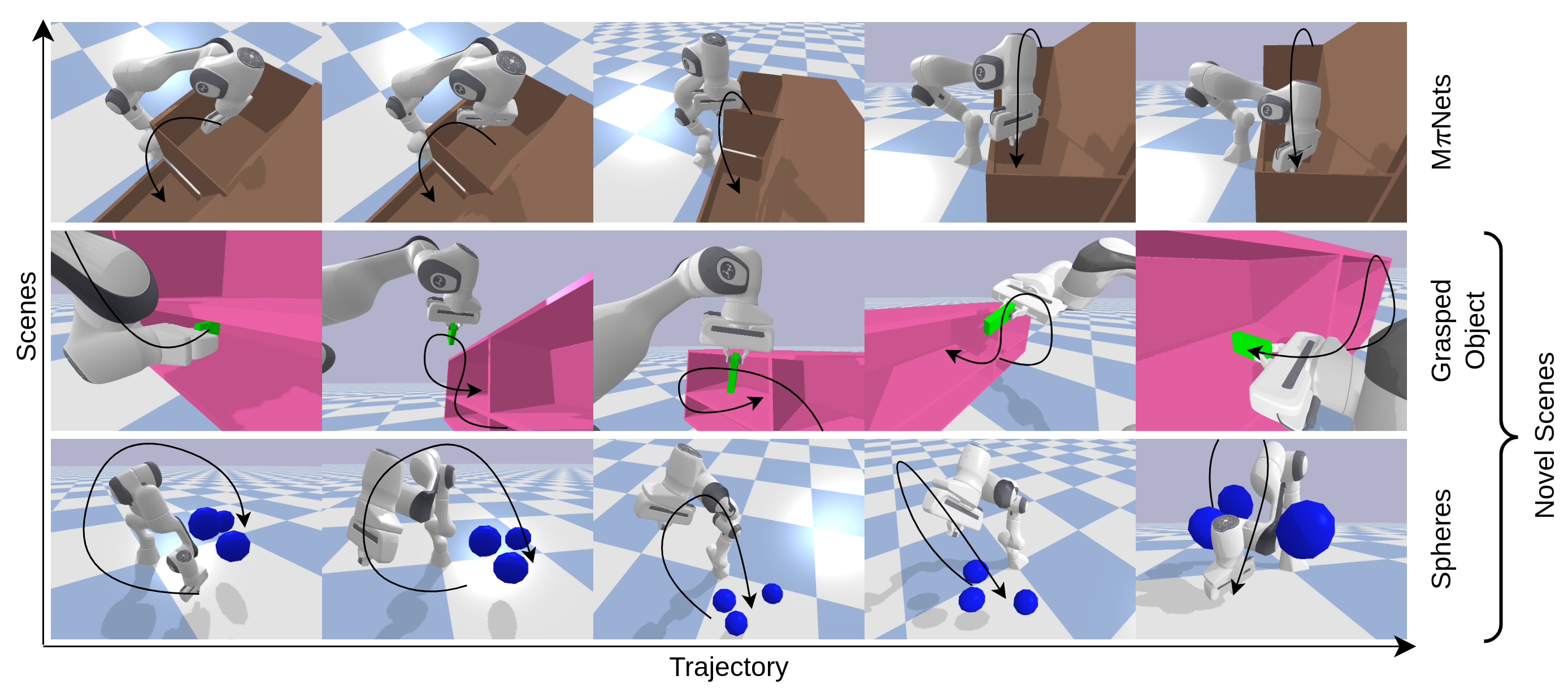}
    \vspace{-8px}
    \caption{\small \textbf{Qualitative Results}: From top-to-bottom: M$\pi$Nets scene (dresser) and out-of-distribution scenes: merged cubby with handheld cuboid and a collision spheres scene.}
    \label{fig:qual}
    \vspace{-10px}
\end{figure*}

The \textbf{test} dataset consists of three datasets: (1) Global Solvable Problem (\textit{\textbf{global}}): consists of 1800 scenes that could be solved by only the Global Planner, that is, Global Planner could generate a valid collision-free trajectory for these scenes. (2) Hybrid Solvable Problems (\textit{\textbf{hybrid}}): consists of 1800 scenes solvable by only the Hybrid Planner. (3) Both Solvable Problem (\textit{\textbf{both}}): a set of 1800 scenes that could be solved by both, Global and Hybrid Planners. More information regarding data collection can be found in \cite{fishman2023motion}.

\textbf{Training and architectural details: }Similar to M$\pi$Nets, we train two different planners; one on \textit{global} and another on \textit{hybrid} data.
Each denoiser is modeled as a temporal UNet similar to~\cite{janner2022planning}, and is trained for 20k steps on a 2080 GeForce GTX GPU for approximately 9 hours. Each \textbf{scene} consists of obstacle configurations, a start position (in joint configuration), a goal position (in joint configuration during training and in end-effector space during test). We use inverse kinematics to compute the joint configuration from the end-effector position. Each \textbf{trajectory} consists of 50 waypoints, including the initial and final joint configurations. For our experiments, we use an ensemble of $12$ cost functions, $5$ with intersection and $7$ with swept volume cost. However, one can define their own number and type of cost functions. 



\textbf{Baselines:} 
We compare our framework against different types of SOTA planners, two behavior cloning-based planners (M$\pi$Nets~\cite{fishman2023motion}, MPNets~\cite{8793889_qureshi}) and two stochastic optimization-based planners (G. Fabrics~\cite{xie2021geometric}, STORM~\cite{bhardwaj2022storm}), one optimization-based planner with quintic-spline initialization (CHOMP~\cite{5152817_ratliff}), and a set of sampling-based planners through OMPL~\cite{6377468_sucan} on \textit{global}, \textit{hybrid}, and \textit{both} test sets. 


\textbf{Metrics:} We define success rate (SR $\%$) as the percentage of problems that are successfully solved by the given planner. We say that a problem is successfully solved if the planner can generate a trajectory that avoids all collisions in order to reach the goal. 
M$\pi$Nets doesn't release the full global data, therefore the metrics are taken from the paper directly. 

\subsection{Performance on M$\pi$Nets Dataset}

\begin{table}[t]
\begin{center}
\adjustbox{max width=\linewidth}{
\centering
\begin{tabular}{c | c c c c c c c c c c}
\Xhline{3\arrayrulewidth} 
Test & CHOMP & OMPL & G. Fabrics & STORM & EDMP \\

\Xhline{2\arrayrulewidth}
Global & 26.67 & 37.27 & 38.44 & 50.22 &  \textbf{75.93} \\
Hybrid & 31.61 & 40.37 & 59.33 & 74.5 &  \textbf{86.13} \\
Both & 32.2 & 42.6 & 60.06 & 76 &  \textbf{85.06} \\
\Xhline{3\arrayrulewidth}
\end{tabular}}
\end{center}
\vspace{-8px}
\captionsetup{textfont=normalfont}
\caption{\small Comparison of EDMP (trained on Hybrid~\cite{fishman2023motion} dataset) against classical motion planners in terms of success rate (\%, $\uparrow$).}
\label{tab:comp_with_classic}
\vspace{-8px}
\end{table}

\begin{table}[t]

\centering
\begin{tabular}{c|cc|ccc}
\Xhline{3\arrayrulewidth}
 & \multicolumn{2}{c|}{Global-trained Planners} & \multicolumn{3}{c}{Hybrid-trained Planners} \\
Test & \makecell{M$\pi$Nets} & \makecell{EDMP} & \makecell{MPNets} & M$\pi$Nets & EDMP \\
\Xhline{2\arrayrulewidth}
Global & \textbf{75.06} & 71.67 & 41.33 & 75.78 & \textbf{75.93} \\
Hybrid & 80.39 & \textbf{82.84} & 65.28 & \textbf{95.33} & 86.13 \\
Both & 82.78 & \textbf{82.79} & 67.67 & \textbf{95.06} & 85.06 \\
\Xhline{3\arrayrulewidth}
\end{tabular}
\captionsetup{textfont=normalfont}
\caption{\small Comparison against deep-learning based methods in terms of success rate (\%, $\uparrow$). EDMP is only marginally influenced by the training dataset. This is because the major performance gain is a result of the ensemble-of-costs-guide directly at inferences common across all the settings.} 
\label{tab:behaviorCloning}
\vspace{-13px}
\end{table}

Table~\ref{tab:comp_with_classic} and \ref{tab:behaviorCloning} presents the success rates of EDMP against the baselines. As shown in Table~\ref{tab:comp_with_classic}, EDMP outperforms all of the classical planners significantly and MPNets (Table~\ref{tab:behaviorCloning}). Although M$\pi$Nets is better than EDMP in many of the settings, it is important to note that our method is agnostic of the variations in the dataset, whereas the performance of methods like M$\pi$Nets is dependent on the dataset it is behaviorally cloned on. This is because our main performance improvement comes from the ensemble of cost functions that is applied directly at the time of inference and is common across all of the settings (\textit{global}, \textit{hybrid}, and \textit{both}). Moreover, the prior encodes kinematic constraints, which can be learned even from much simpler training datasets like \textit{global} instead of \textit{hybrid}. Moreover, unlike M$\pi$Nets, EDMP generalizes to out-of-distribution scenes (Section~\ref{sec:ood}) and generates multimodal trajectories (Section~\ref{sec:mm}).

\subsection{Generalization to Out-of-distribution Scenes}
\label{sec:ood}
We test EDMP's performance on scenes outside of the training distribution. First, we make the manipulator hold arbitrary objects and see if EDMP can find a \textit{valid collision-free} trajectory for the manipulator along with the object. As can be seen in Figure~\ref{fig:qual}, EDMP works out of the box for such scenes. Deep-learning-based approaches like M$\pi$Nets and MPNets require retraining or specialized training datasets to take objects into account. In our case, we treat the object as just another link (as explained in Section ~\ref{sec:cost_guidance}) with fixed joints and adding an object-environment collision-cost to the overall cost function (Equation~\ref{algo:reverse_diff}). 

We also test EDMP by generating a scene with collision spheres (Figure~\ref{fig:qual}). Such a scene is very different from the types of scenes the training trajectories were generated on. Even in this highly challenging setup, EDMP succeeds in 8 out of 10 evaluated scenes with spheres generated at random spatial positions, indicating that EDMP has learned a generic prior that is highly adaptable to diverse scenes. 

\subsection{Multimodality} 
\label{sec:mm}
Multimodal trajectories can enable one to achieve the same goal in multiple different ways. This is much closer to how humans plan -- out of many different ways to execute a task; we often pick the one that is most optimal for the \textit{current} situation. Based on the concept of diffusion, we inherently support such multimodal trajectories. Moreover, our ensemble-of-costs further improves the diversity in the generated trajectories.

We assess trajectory diversity for 100 random scenes from the M$\pi$Nets validation dataset. We use Average Cosine Similarity Mean (ACSM) metric that calculates the mean of cosine similarities among all pairs of trajectories within a batch to measure the diversity in trajectories. Lower ACSM suggests greater diversity. As shown in Table~\ref{tab:Multimodality_results}, our ensemble-of-costs guidance outperforms the average of 12 costs function when measured individually, demonstrating the advantage of multiple cost-guidance over a single one.

\begin{table}[t]
\centering
\begin{tabular}{ccc}  
\Xhline{3\arrayrulewidth}
\textbf{Method} & \makecell{\textbf{SR}($\uparrow$)} & \makecell{\textbf{ACSM}($\downarrow$)} 
\\
\Xhline{2\arrayrulewidth}
\makecell{Avg. of single guides} & 79.25 & 0.981 
\\
EDMP & \textbf{89.30} & \textbf{0.892} 
\\
\Xhline{3\arrayrulewidth}
\end{tabular}
\vspace{-3px}
\captionsetup{textfont=normalfont}
\caption{\small Comparison of EDMP against the individual guides on 100 M$\pi$Net validation scenes.}
\vspace{-5px}
\label{tab:Multimodality_results}
\end{table}

\section{ABLATION STUDIES}
\label{sec:ablations}

\begin{table}[t]
\centering
\begin{tabular}{c c c c c c}
\Xhline{3\arrayrulewidth}
Condition & \makecell{AR($\downarrow$)} & MRESG($\downarrow$) & RF2W($\downarrow$) & RL2W($\downarrow$) \\
\Xhline{2\arrayrulewidth}
\makecell{w/cond} & \textbf{0.0589} & \textbf{0.0821} & \textbf{0.0266} & \textbf{0.0266} \\
\makecell{wo/cond} & 0.1121 & 0.3171 & 0.6459 & 0.4960 \\
\Xhline{3\arrayrulewidth}
\end{tabular}
\vspace{-3px}
\captionsetup{textfont=normalfont}
\caption{\small \textbf{Ablation on start-goal conditioned trajectory}: As we generate the trajectory through the denoising process conditioned on the start and goal, the trajectories are smoother. The metrics are defined in Section~\ref{sec:ablations}}.
\label{tab:training_conditioning}
\vspace{-8px}
\end{table}

\begin{figure}[t!]
    \centering
    \includegraphics[width=0.95\linewidth]{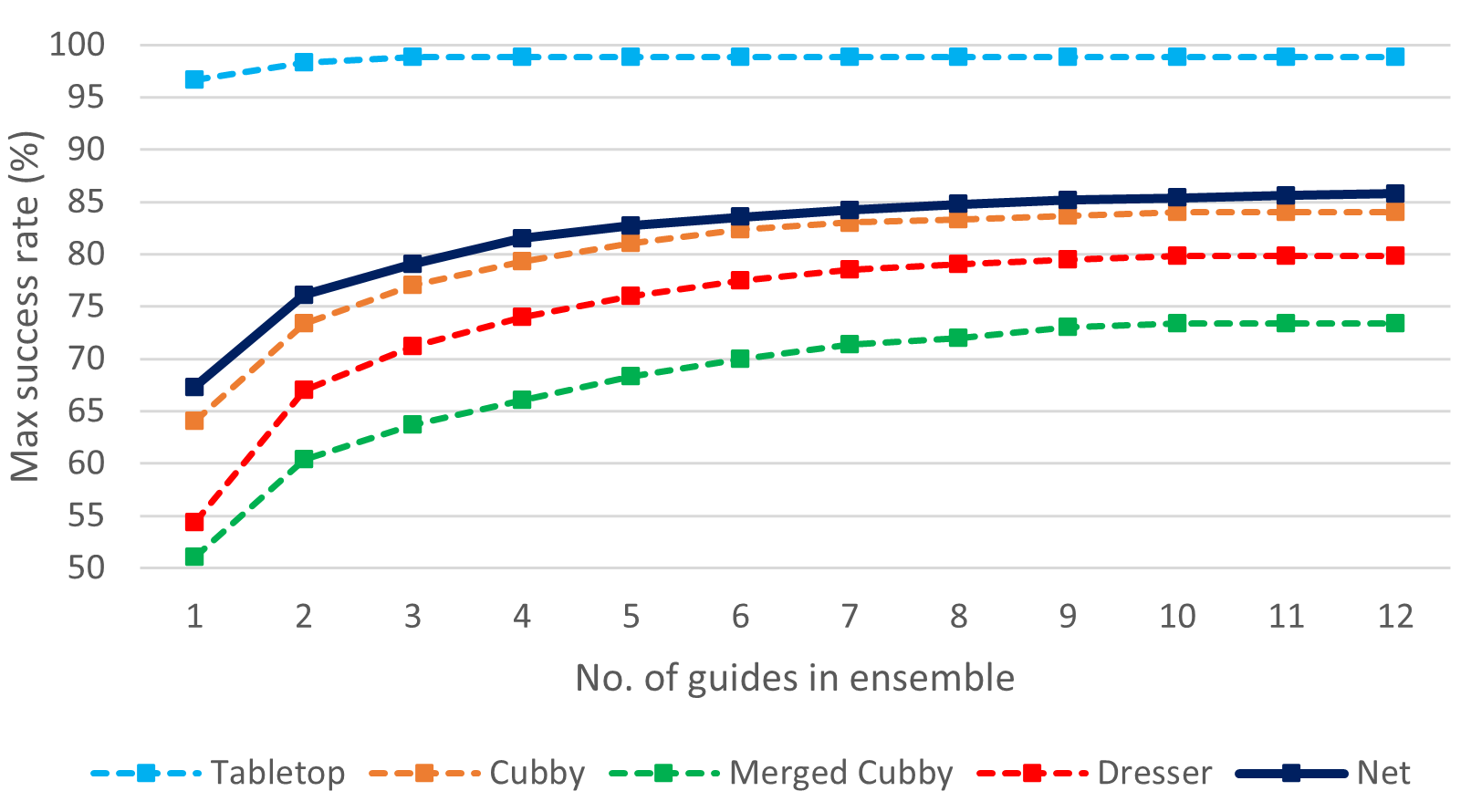}
    \vspace{-3px}
    \caption{\small \textbf{Effect of adding guides.} There is an asymptotic rise in the overall Success Rates with the number of guides demonstrated across diverse scenes - tabletop, cubby, merged cubby, and dresser.}
    \label{fig:guide_char1}
    \vspace{-10px}
\end{figure}

\begin{figure}[t!]
    \centering
    \includegraphics[width=0.95\linewidth, height=140px]{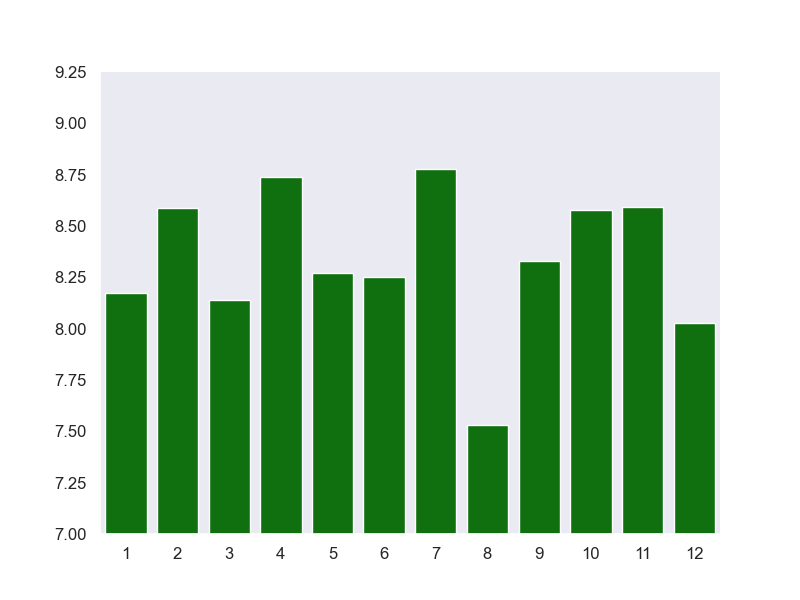}
    \vspace{-10px}
    \caption{\small \textbf{Contribution of each guide.} x-axis and y-axis denote the guide index and contribution (\%) of each guide to overall Success Rate. The contribution of a guide is proportional to the likelihood that a successful path was generated by it.} 
    \label{fig:guide_char2}
    \vspace{-15px}
\end{figure}

\textbf{Ensemble-of-costs.} In this section we show the effect of ensemble-of-costs in Figure~\ref{fig:guide_char1} and \ref{fig:guide_char2}. As can be seen in Figure~\ref{fig:guide_char1}, as the number of guides increase, the overall Success Rates increases. This is because diverse cost functions capture different parameters and aspects across scenes, as explained in Section~\ref{sec:ensemble}. Although these are specially designed cost functions, the overall idea of combining multiple cost functions to solve diverse scenes can be applied in any domain using custom cost functions. Figure~\ref{fig:guide_char2} further emphasizes the contribution of each guide to the overall success rate.

\textbf{Conditioning with start and goal.}
One primary condition of generating a trajectory is avoiding any abrupt manipulator motion that could cause serious damage in the real-world. We evaluate the effect of conditioning the trajectory with the start and goal. We defined Roughness as the L2 norm between adjacent waypoints. To evaluate this, we consider four metrics - Average Roughness (AR), Max Roughness Excluding Start and Goal (MRESG), Roughness between first two waypoints (RF2W), and Roughness between last two waypoints (RL2W).  
As shown in Table~\ref{tab:training_conditioning}, conditional training generates significantly smoother trajectories. 

\section{CONCLUSION \& FUTURE WORK}
In this paper, we propose diffusion models as a mechanism to learn a prior over kinematically valid trajectories complemented by scene-specific guidance through gradient-based cost functions directly at inference.
We introduce an ensemble of cost-functions 
that capture scene-specific nuances across diverse scenes and improve the generated trajectories significantly.  
Our empirical results concretely point towards our approach's ability to effectively exploit the strengths of classical motion planners and the recent deep-learning-based planners,
by showcasing remarkable generalization capability across diverse scenes, such as object manipulation, without any object-specific retraining.
Moreover, EDMP is capable of generating diverse multimodal trajectories 
for a given scene, facilitating optimal trajectory selection based on specific downstream parameters. We believe that our research provides significant strides towards more generalizable planning with broad applications to robotic manipulation. 

\textbf{Future work. }Despite showcasing impressive performance and generalization, EDMP still relies on a set of handcrafted cost functions to capture diversity across the scenes. Automating the design of cost-functions based on scene diversity could be an interesting future work.


\section*{ACKNOWLEDGMENT}
We would like to thank Adam Fishman for assisting with M$\pi$Nets and providing valuable insights into collision checking and benchmarking.

\printbibliography

@ARTICLE{4082128,
  author={Hart, Peter E. and Nilsson, Nils J. and Raphael, Bertram},
  journal={IEEE Transactions on Systems Science and Cybernetics}, 
  title={A Formal Basis for the Heuristic Determination of Minimum Cost Paths}, 
  year={1968},
  volume={4},
  number={2},
  pages={100-107},
  doi={10.1109/TSSC.1968.300136}}

@article{agarwal2022approaches,
  title={Approaches and challenges in robotic perception for table-top rearrangement and planning},
  author={Agarwal, Aditya and Sen, Bipasha and Narayanan V, Shankara and Mandadi, Vishal Reddy and Bhowmick, Brojeshwar and Krishna, K Madhava},
  journal={arXiv preprint arXiv:2205.04090},
  year={2022}
}

@inproceedings{conf/nips/LikhachevGT03,
  author = {Likhachev, Maxim and Gordon, Geoffrey J. and Thrun, Sebastian},
  booktitle = {NIPS},
  crossref = {conf/nips/2003},
  editor = {Thrun, Sebastian and Saul, Lawrence K. and Schölkopf, Bernhard},
  ee = {http://papers.nips.cc/paper/2382-ara-anytime-a-with-provable-bounds-on-sub-optimality},
  isbn = {0-262-20152-6},
  pages = {767-774},
  publisher = {MIT Press},
  title = {ARA*: Anytime A* with Provable Bounds on Sub-Optimality.},
  url = {http://dblp.uni-trier.de/db/conf/nips/nips2003.html#LikhachevGT03},
  year = 2003
}

@inproceedings{inproceedings,
author = {Likhachev, Maxim and Ferguson, David and Gordon, Geoffrey and Stentz, Anthony and Thrun, Sebastian},
year = {2005},
month = {01},
pages = {262-271},
title = {Anytime Dynamic A*: An Anytime, Replanning Algorithm.},
journal = {Proceedings of the International Conference on Automated Planning and Scheduling (ICAPS)}
}

@article{LaValle1998RapidlyexploringRT,
  title={Rapidly-exploring random trees : a new tool for path planning},
  author={Steven M. LaValle},
  journal={The annual research report},
  year={1998},
  url={https://api.semanticscholar.org/CorpusID:14744621}
}

@article{Kuffner2000RRTconnectAE,
  title={RRT-connect: An efficient approach to single-query path planning},
  author={James J. Kuffner and Steven M. LaValle},
  journal={Proceedings 2000 ICRA. Millennium Conference. IEEE International Conference on Robotics and Automation. Symposia Proceedings (Cat. No.00CH37065)},
  year={2000},
  volume={2},
  pages={995-1001 vol.2},
  url={https://api.semanticscholar.org/CorpusID:17124403}
}

@article{Williams2015ModelPP,
  title={Model Predictive Path Integral Control using Covariance Variable Importance Sampling},
  author={Grady Williams and Andrew Aldrich and Evangelos A. Theodorou},
  journal={ArXiv},
  year={2015},
  volume={abs/1509.01149},
  url={https://api.semanticscholar.org/CorpusID:14146342}
}

@article{Jankowski2022VPSTOVS,
  title={VP-STO: Via-point-based Stochastic Trajectory Optimization for Reactive Robot Behavior},
  author={Julius Jankowski and Lara Bruderm{\"u}ller and Nick Hawes and Sylvain Calinon},
  journal={2023 IEEE International Conference on Robotics and Automation (ICRA)},
  year={2022},
  pages={10125-10131},
  url={https://api.semanticscholar.org/CorpusID:252780630}
}

@INPROCEEDINGS{7487277,
  author={Williams, Grady and Drews, Paul and Goldfain, Brian and Rehg, James M. and Theodorou, Evangelos A.},
  booktitle={2016 IEEE International Conference on Robotics and Automation (ICRA)}, 
  title={Aggressive driving with model predictive path integral control}, 
  year={2016},
  volume={},
  number={},
  pages={1433-1440},
  doi={10.1109/ICRA.2016.7487277}}

@inproceedings{bhardwaj2022storm,
  title={Storm: An integrated framework for fast joint-space model-predictive control for reactive manipulation},
  author={Bhardwaj, Mohak and Sundaralingam, Balakumar and Mousavian, Arsalan and Ratliff, Nathan D and Fox, Dieter and Ramos, Fabio and Boots, Byron},
  booktitle={Conference on Robot Learning},
  pages={750--759},
  year={2022},
  organization={PMLR}
}

@inproceedings{inproceedings_bangura,
author = {Bangura, Moses},
year = {2014},
month = {08},
pages = {11773-11780},
title = {Real-Time Model Predictive Control for Quadrotors},
volume = {47},
journal = {IFAC Proceedings Volumes},
doi = {10.3182/20140824-6-ZA-1003.00203}
}

@INPROCEEDINGS{7029990_erez,
  author={Erez, Tom and Lowrey, Kendall and Tassa, Yuval and Kumar, Vikash and Kolev, Svetoslav and Todorov, Emanuel},
  booktitle={2013 13th IEEE-RAS International Conference on Humanoid Robots (Humanoids)}, 
  title={An integrated system for real-time model predictive control of humanoid robots}, 
  year={2013},
  volume={},
  number={},
  pages={292-299},
  doi={10.1109/HUMANOIDS.2013.7029990}}

@INPROCEEDINGS{5152817_ratliff,
  author={Ratliff, Nathan and Zucker, Matt and Bagnell, J. Andrew and Srinivasa, Siddhartha},
  booktitle={2009 IEEE International Conference on Robotics and Automation}, 
  title={CHOMP: Gradient optimization techniques for efficient motion planning}, 
  year={2009},
  volume={},
  number={},
  pages={489-494},
  doi={10.1109/ROBOT.2009.5152817}}

@inproceedings{inproceedings_schulman_ho_jonathan_lee,
author = {Schulman, John and Ho, Jonathan and Lee, Alex and Awwal, Ibrahim and Bradlow, Henry and Abbeel, Pieter},
year = {2013},
month = {06},
pages = {},
title = {Finding Locally Optimal, Collision-Free Trajectories with Sequential Convex Optimization},
doi = {10.15607/RSS.2013.IX.031}
}

@ARTICLE{6377468_sucan,
  author={Sucan, Ioan A. and Moll, Mark and Kavraki, Lydia E.},
  journal={IEEE Robotics \& Automation Magazine}, 
  title={The Open Motion Planning Library}, 
  year={2012},
  volume={19},
  number={4},
  pages={72-82},
  doi={10.1109/MRA.2012.2205651}}

@INPROCEEDINGS{8793889_qureshi,
  author={Qureshi, Ahmed H. and Simeonov, Anthony and Bency, Mayur J. and Yip, Michael C.},
  booktitle={2019 International Conference on Robotics and Automation (ICRA)}, 
  title={Motion Planning Networks}, 
  year={2019},
  volume={},
  number={},
  pages={2118-2124},
  doi={10.1109/ICRA.2019.8793889}}

@inproceedings{fishman2023motion,
  title={Motion policy networks},
  author={Fishman, Adam and Murali, Adithyavairavan and Eppner, Clemens and Peele, Bryan and Boots, Byron and Fox, Dieter},
  booktitle={Conference on Robot Learning},
  pages={967--977},
  year={2023},
  organization={PMLR}
}

@inproceedings{danielczuk2021object,
  title={Object rearrangement using learned implicit collision functions},
  author={Danielczuk, Michael and Mousavian, Arsalan and Eppner, Clemens and Fox, Dieter},
  booktitle={2021 IEEE International Conference on Robotics and Automation (ICRA)},
  pages={6010--6017},
  year={2021},
  organization={IEEE}
}

@article{osa2018algorithmic,
  title={An algorithmic perspective on imitation learning},
  author={Osa, Takayuki and Pajarinen, Joni and Neumann, Gerhard and Bagnell, J Andrew and Abbeel, Pieter and Peters, Jan and others},
  journal={Foundations and Trends{\textregistered} in Robotics},
  volume={7},
  number={1-2},
  pages={1--179},
  year={2018},
  publisher={Now Publishers, Inc.}
}

@article{pearce2023imitating,
  title={Imitating human behaviour with diffusion models},
  author={Pearce, Tim and Rashid, Tabish and Kanervisto, Anssi and Bignell, Dave and Sun, Mingfei and Georgescu, Raluca and Macua, Sergio Valcarcel and Tan, Shan Zheng and Momennejad, Ida and Hofmann, Katja and others},
  journal={arXiv preprint arXiv:2301.10677},
  year={2023}
}

@article{zhao2023learning_aloha,
  title={Learning fine-grained bimanual manipulation with low-cost hardware},
  author={Zhao, Tony Z and Kumar, Vikash and Levine, Sergey and Finn, Chelsea},
  journal={arXiv preprint arXiv:2304.13705},
  year={2023}
}

@misc{dhariwal2021diffusion,
      title={Diffusion Models Beat GANs on Image Synthesis}, 
      author={Prafulla Dhariwal and Alex Nichol},
      year={2021},
      eprint={2105.05233},
      archivePrefix={arXiv},
      primaryClass={cs.LG}
}

@misc{goodfellow2014generative,
      title={Generative Adversarial Networks}, 
      author={Ian J. Goodfellow and Jean Pouget-Abadie and Mehdi Mirza and Bing Xu and David Warde-Farley and Sherjil Ozair and Aaron Courville and Yoshua Bengio},
      year={2014},
      eprint={1406.2661},
      archivePrefix={arXiv},
      primaryClass={stat.ML}
}

@misc{kingma2022autoencoding,
      title={Auto-Encoding Variational Bayes}, 
      author={Diederik P Kingma and Max Welling},
      year={2022},
      eprint={1312.6114},
      archivePrefix={arXiv},
      primaryClass={stat.ML}
}

@ARTICLE{2083,
  author={Gilbert, E.G. and Johnson, D.W. and Keerthi, S.S.},
  journal={IEEE Journal on Robotics and Automation}, 
  title={A fast procedure for computing the distance between complex objects in three-dimensional space}, 
  year={1988},
  volume={4},
  number={2},
  pages={193-203},
  doi={10.1109/56.2083}}

@article{janner2022planning,
  title={Planning with diffusion for flexible behavior synthesis},
  author={Janner, Michael and Du, Yilun and Tenenbaum, Joshua B and Levine, Sergey},
  journal={arXiv preprint arXiv:2205.09991},
  year={2022}
}

@article{ajay2022conditional,
  title={Is conditional generative modeling all you need for decision-making?},
  author={Ajay, Anurag and Du, Yilun and Gupta, Abhi and Tenenbaum, Joshua and Jaakkola, Tommi and Agrawal, Pulkit},
  journal={arXiv preprint arXiv:2211.15657},
  year={2022}
}

@misc{liang2023adaptdiffuser,
      title={AdaptDiffuser: Diffusion Models as Adaptive Self-evolving Planners}, 
      author={Zhixuan Liang and Yao Mu and Mingyu Ding and Fei Ni and Masayoshi Tomizuka and Ping Luo},
      year={2023},
      eprint={2302.01877},
      archivePrefix={arXiv},
      primaryClass={cs.LG}
}

@article{chi2023diffusion,
  title={Diffusion policy: Visuomotor policy learning via action diffusion},
  author={Chi, Cheng and Feng, Siyuan and Du, Yilun and Xu, Zhenjia and Cousineau, Eric and Burchfiel, Benjamin and Song, Shuran},
  journal={arXiv preprint arXiv:2303.04137},
  year={2023}
}

@article{carvalho2023motion,
  title={Motion Planning Diffusion: Learning and Planning of Robot Motions with Diffusion Models},
  author={Carvalho, Joao and Le, An T and Baierl, Mark and Koert, Dorothea and Peters, Jan},
  journal={arXiv preprint arXiv:2308.01557},
  year={2023}
}

@inproceedings{sohl2015deep,
  title={Deep unsupervised learning using nonequilibrium thermodynamics},
  author={Sohl-Dickstein, Jascha and Weiss, Eric and Maheswaranathan, Niru and Ganguli, Surya},
  booktitle={International conference on machine learning},
  pages={2256--2265},
  year={2015},
  organization={PMLR}
}

@article{ho2020denoising,
  title={Denoising diffusion probabilistic models},
  author={Ho, Jonathan and Jain, Ajay and Abbeel, Pieter},
  journal={Advances in neural information processing systems},
  volume={33},
  pages={6840--6851},
  year={2020}
}

@article{article,
author = {Bergen, Gino},
year = {2001},
month = {01},
pages = {},
title = {Proximity queries and penetration depth computation on 3d game objects}
}

@inproceedings{strub2020adaptively,
  title={Adaptively Informed Trees (AIT): Fast Asymptotically Optimal Path Planning through Adaptive Heuristics},
  author={Strub, Marlin P and Gammell, Jonathan D},
  booktitle={2020 IEEE International Conference on Robotics and Automation (ICRA)},
  pages={3191--3198},
  year={2020},
  organization={IEEE}
}

@misc{xie2021geometric,
      title={Geometric Fabrics for the Acceleration-based Design of Robotic Motion}, 
      author={Mandy Xie and Karl Van Wyk and Anqi Li and Muhammad Asif Rana and Qian Wan and Dieter Fox and Byron Boots and Nathan Ratliff},
      year={2021},
      eprint={2010.14750},
      archivePrefix={arXiv},
      primaryClass={cs.RO}
}

@MISC{coumans2021,
author =   {Erwin Coumans and Yunfei Bai},
title =    {PyBullet, a Python module for physics simulation for games, robotics and machine learning},
howpublished = {\url{http://pybullet.org}},
year = {2016--2021}
}

@inproceedings{dalle,
  title={Zero-shot text-to-image generation},
  author={Ramesh, Aditya and Pavlov, Mikhail and Goh, Gabriel and Gray, Scott and Voss, Chelsea and Radford, Alec and Chen, Mark and Sutskever, Ilya},
  booktitle={International Conference on Machine Learning},
  pages={8821--8831},
  year={2021},
  organization={PMLR}
}

@article{dalle2,
  title={Hierarchical text-conditional image generation with clip latents},
  author={Ramesh, Aditya and Dhariwal, Prafulla and Nichol, Alex and Chu, Casey and Chen, Mark},
  journal={arXiv preprint arXiv:2204.06125},
  volume={1},
  number={2},
  pages={3},
  year={2022}
}

@article{imagen,
  title={Photorealistic text-to-image diffusion models with deep language understanding},
  author={Saharia, Chitwan and Chan, William and Saxena, Saurabh and Li, Lala and Whang, Jay and Denton, Emily L and Ghasemipour, Kamyar and Gontijo Lopes, Raphael and Karagol Ayan, Burcu and Salimans, Tim and others},
  journal={Advances in Neural Information Processing Systems},
  volume={35},
  pages={36479--36494},
  year={2022}
}

@inproceedings{audio_video_diffusion,
  title={Mm-diffusion: Learning multi-modal diffusion models for joint audio and video generation},
  author={Ruan, Ludan and Ma, Yiyang and Yang, Huan and He, Huiguo and Liu, Bei and Fu, Jianlong and Yuan, Nicholas Jing and Jin, Qin and Guo, Baining},
  booktitle={Proceedings of the IEEE/CVF Conference on Computer Vision and Pattern Recognition},
  pages={10219--10228},
  year={2023}
}
\end{document}